\documentclass[conference]{IEEEtran}
\usepackage{multicol}
\usepackage{amsmath}
\usepackage{algorithm}
\usepackage{graphicx}
\usepackage[noend]{algpseudocode}
\usepackage{url}

\begin{document}






%

\title{Distributed Training of Deep Neural Networks: Theoretical and Practical Limits of Parallel Scalability.}
%
%
%
%
%

\author{\IEEEauthorblockN{Janis Keuper}
\IEEEauthorblockA{Fraunhofer ITWM\\
Competence Center High Performance Computing\\
Kaiserslautern, Germany\\
Email: janis.keuper@itwm.fhg.de}
\and
\IEEEauthorblockN{Franz-Josef Pfreundt}
\IEEEauthorblockA{Fraunhofer ITWM\\
Competence Center High Performance Computing\\
Kaiserslautern, Germany\\
Email: franz-josef.pfreundt@itwm.fhg.de}

}

\maketitle
\begin{abstract}
This paper presents a theoretical analysis and practical evaluation of the main 
bottlenecks towards a scalable distributed solution for the training of
Deep Neural Networks (DNNs). The presented results show, that the current state 
of the art approach, using data-parallelized Stochastic Gradient Descent (SGD), is
quickly turning into a vastly communication bound problem. In addition, we present simple but 
fixed theoretic constraints, preventing effective scaling of DNN training beyond 
only a few dozen nodes. This leads to poor scalability 
of DNN training in most practical scenarios.        
\end{abstract}

%
%



\section{Introduction}
\begin{figure}[t!]
\includegraphics[width=\linewidth]{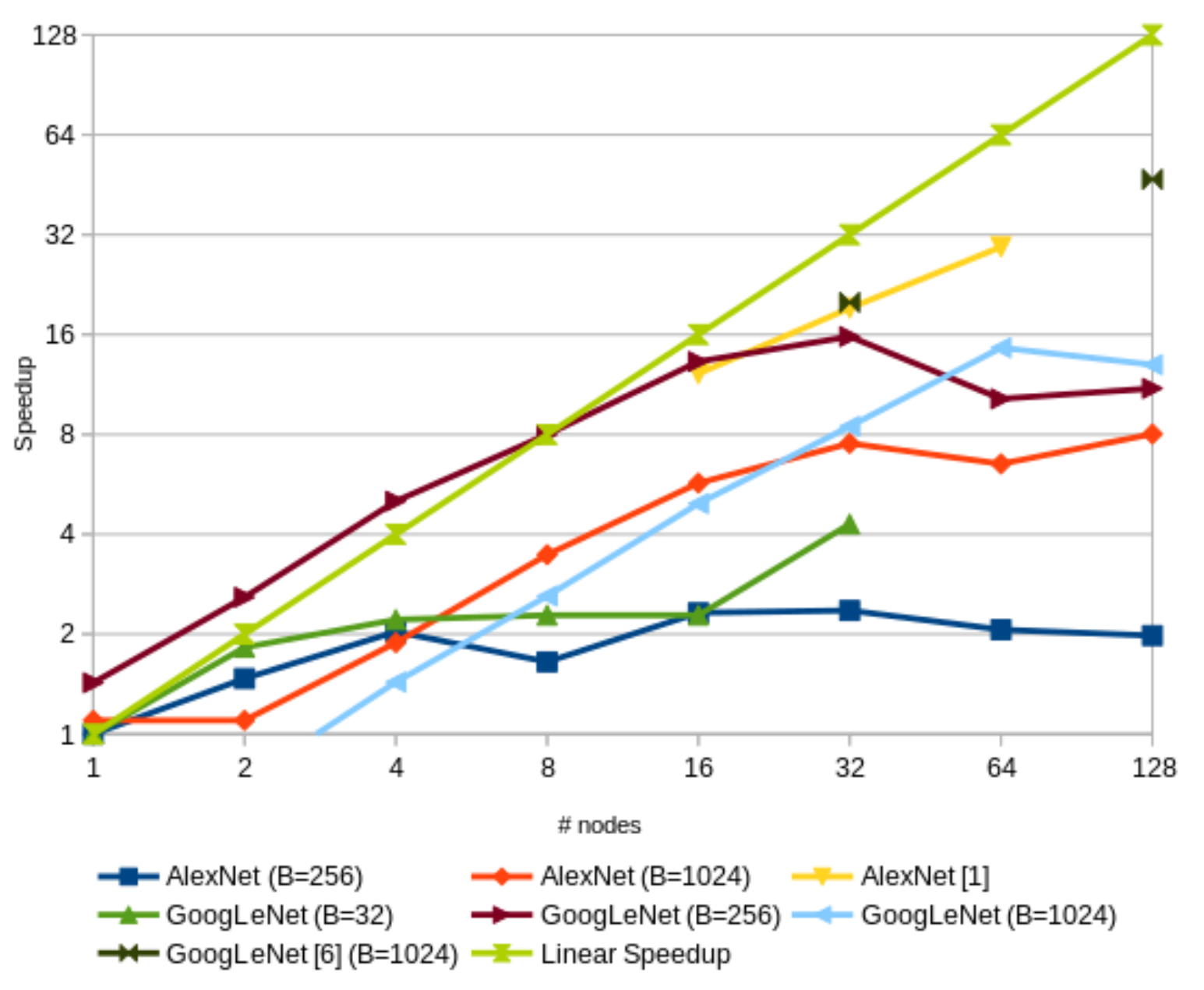}
\caption{Experimental evaluation of DNN training scalability (strong scaling) for different DNNs with varying global batch sizes $B$. 
Results from an "out of the box" installation of {\it IntelCaffe} on a common HPC system (Details are given in section \ref{sec:exp_setup})  \label{fig:Baseline}}
\end{figure}

\noindent The tremendous success of Deep Neural Networks (DNNs) \cite{schmidhuber2015deep, lecun2015deep}
in a wide range of practically relevant
applications has triggered a race to build larger and larger DNNs \cite{simonyan2014very}, 
which need to be trained with 
more and more data, to solve learning problems in fast extending fields of applications. 
\begin{table}[ht]
\centering
\begin{tabular}{l|cccc}
 & CPU & K80 & TitanX & KNL \\
\hline
AlexNet:& & &  & \\
time per iteration & 2s & 0.9s & 0.2s \cite{cnnbench} & 0.6s \\
time till convergence & 250h & 112h & 25h \cite{cnnbench}& 75h \\
GoogLeNet:& & &  & \\
time per iteration & 1.3s& 0.36s & - & 0.32s \\
time till convergence & 361h & 100h & -  & 89h  \\
\end{tabular}
\caption{Approximate computation times for AlexNet with batch size $B=256$ and 450k iterations and  GoogLeNet with $B=32$ and 1000k iterations. KNL (Xeon Phi ``Knights Landing'') results with MKL17. TitanX with Pascal
GPU. See section \ref{sec:hardware}. \label{tab:compute}}
\end{table}
However,
training DNNs is a compute and data intensive task: current models take several ExaFLOP to
compute, while processing hundreds of petabyte of data \cite{simonyan2014very}.
Table \ref{tab:compute} gives an impression
of the compute complexity and shows,
that even the latest compute hardware will take days to train the medium sized benchmark networks
used in our experiments.
While a parallelization of the training problem over up to 8 GPUs hosted in a single
compute node can be considered to be the current state of the art, available distributed approaches 
\cite{dean2012large, ma2016theano, intelcaffe, abadi2016tensorflow, DBLP:journals/corr/IandolaAMK15}
yield disappointing results \cite{seide2014parallelizability} in terms of scalability and efficiency. 
Figure \ref{fig:Baseline}
shows representative experimental evaluations, where strong scaling is stalling after only a few dozen
nodes.
\noindent In this paper, we investigate the theoretical and practical constraints preventing better scalability, 
namely
model distribution overheads (in section \ref{sec:dist_overhead}), data-parallelized matrix multiplication 
(section \ref{sec:matmul}) and training data distribution (section \ref{sec:distdata}). 

\begin{figure*}[ht]
\includegraphics[width=\linewidth]{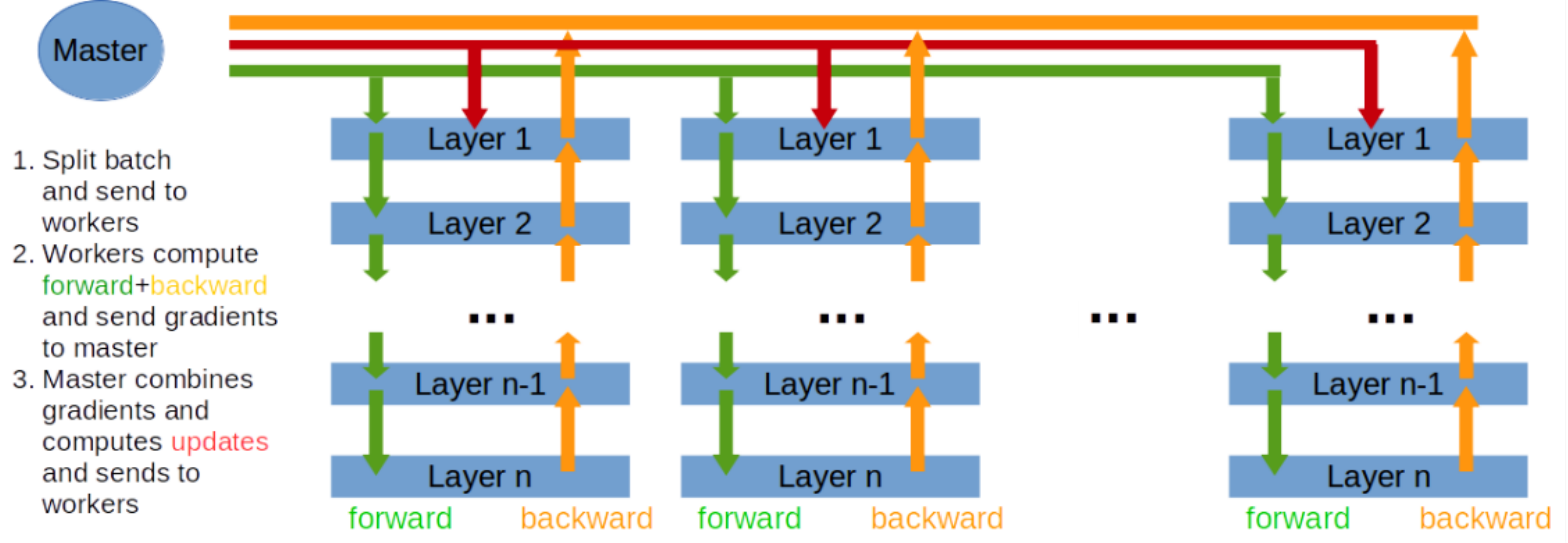}
\caption{Schematic overview of a distributed SGD implementation of the {\it Backpropagation Algorithm}.  \label{fig:scheme}}
\end{figure*}

\subsection{Stochastic Gradient Descent}
\noindent Deep Neural Networks are trained using the {\it Backpropagation Algorithm} 
\cite{ngiam2011optimization}. Numerically, this is formulated
as a highly non-convex optimization problem in a very high dimensional space, which is typically solved
via Stochastic Gradient Descent (SGD\footnote{Usually, SGD with additional 2nd order terms (moments) 
are used, but this has no impact on the parallelization.}) \cite{bottou2010large}. 
SGD, using  moderate mini-batch sizes $B$, provides 
stable convergence at fair computational costs on a single node. However, it is  very hard to parallelize. 
This is due to the inherently sequential nature of the algorithm, shown in equation \ref{eq:sgd} and 
algorithm \ref{algo_mini}.

\begin{equation}
w_{t+1} \leftarrow w_{t} - \epsilon\partial_wx_j(w_{t}),
\label{eq:sgd}
\end{equation}
where $w_{t}$ represents the current sate (e.g. the weights at the neurons), $\epsilon$ defines the step-size
and $\partial_wx_j(w_{t})$ is computed from a given loss-function over the forward results of a small set 
of training samples 
(called {\it mini-batch} and the given training labels).
In fact, there are only two ways to speedup SGD: (I) computing updates ${\bf\Delta}w$ faster
and (II) making larger update steps $\epsilon$. 
While (I) is hard to achieve in a distributed setting given already low 
compute times ($<1$s) per iteration, (II) is bound by the 
difficult topologies of the non-convex problems, causing SGD to diverge easily.\\             

\subsubsection{Parallelizing SGD}
\label{sec:PSGD}
\noindent Figure \ref{fig:scheme} shows the data-parallel version of SGD \cite{dean2012large}, 
which is commonly used for 
single node multi-GPU and distributed implementations: The global batch of $B$ training samples for the 
current iteration
is split into $n$ equal sized sets of size $b$ (with $b=B/n$) of training samples which are then fed to $n$ 
workers holding synchronous local
copies of the model state. The results (gradients) off all workers are then accumulated and used to update the
model. Hence, the entire approach is implementing a simple {\it map-reduce} scheme.\\
\noindent Notably, this scheme implies two different levels parallelization: the 
data- and task-parallel \cite{krizhevsky2014one} {\it Inner Parallelization}, 
 located at the compute units of the nodes using parallel algorithms to compute the forward and backward  
operations within the layers of the DNN (see section \ref{sec:matmul} for details on the local parallelization 
of layer operations), and the {\it Outer Parallelization} over the distributed batches.   

\begin{algorithm}
\caption{Mini-Batch SGD
with samples $X=\{x_0,\dots,x_m\}$, iterations $T$, step-size
$\epsilon$, batch size $B$ }
\label{algo_mini}
\begin{algorithmic}[1]
\Require{$\epsilon>0$}
\ForAll{$t=0\dots T$ }
\State{\begin{bf}randomly draw\end{bf} batch $M \leftarrow B$ samples from $X$}
\State{\begin{bf}Init\end{bf}${\bf\Delta}w_t=0$ }
\ForAll{$x\in M$ }
\State{\begin{bf}aggregate update\end{bf} ${\bf\Delta}w \leftarrow \partial_wx_j(w_{t})$}
\EndFor
\State{\begin{bf}update\end{bf} $w_{t+1} \leftarrow w_{t} - \epsilon{\bf\Delta}w_t$}

\EndFor
\State{\Return $w_T$}

\end{algorithmic}
\end{algorithm}


\subsection{Experimental Setup \label{sec:exp_setup}}

\subsubsection{Benchmarks}
\noindent We apply two widely used convolutional networks (CNNs), {\it AlexNet} 
\cite{krizhevsky2012imagenet} and
{\it GoogLeNet} \cite{szegedy2015going}, for the benchmarking of our experimental evaluations.
 Both 
neural networks follow different strategies to learn predictive models for the {\it ImageNet}
 \cite{russakovsky2015imagenet} visual recognition challenge. While {\it AlexNet} implements 
a rather shallow network with 3 dominant fully-connected (FC) layers, is {\it GoogLeNet} using a 
very  deep network with many convolutional layers. Table \ref{tab:bench} shows the technical 
details of both networks.   

\begin{table}[ht]
\centering
\begin{tabular}{l|cc}
& AlexNet & GoogLeNet\\
\hline
ExaFLOP to convergence & $\sim$ 0.8 & $\sim$1.1 \\
\# Iterations till convergence& 450k & 1000k \\
Model size @32 bit FP & $\sim$250 MB & $\sim$50 MB \\
Default batch size & 256 & 32 \\
Default step-size & 0.01 & 0.01 \\
\# Layers & 25 & 159 \\
\# Convolutional layers & 5 & 59 \\
\# Fully-connected (FC) layers & 3 & 1\\
\# Weights in FC layers & $\sim$55M & $\sim$1M\\
\end{tabular}
\caption{Properties of the Deep Neural Networks used for the following benchmarks. \label{tab:bench}}
\end{table}

\subsubsection{Software Framework}
\noindent We use the MPI based distributed Version ({\it IntelCaffe}) \cite{intelcaffe} 
of the popular open source 
framework 
{\it Caffe} \cite{jia2014caffe} for our evaluation. {\it IntelCaffe} was built with 
{\it CUDA 7.5} and {\it cuDNN 5} using the latest {\it Intel} compiler, {\it MKL}\footnote{Some CPU experiments used the latest DNN extensions of the {\it MKL17} library which provides special purpose functions for the fast implementation of several layer types like cuDNN for CUDA. } 
and {\it IntelMPI}.\\

\subsubsection{Hardware\label{sec:hardware}}
\noindent All distributed experiments were conducted on a HPC cluster with nodes holding a dual Xeon 
E5-2680 v3 CPU (12 cores @ 2.50GHz), a NVIDIA Tesla K80 GPU and FDR-Infiniband interconnects.

\section{Distribution Overhead}
\label{sec:dist_overhead}
\noindent The parallelization of DNN training via SGD (as shown in algorithm \ref{algo_mini} and 
figure \ref{fig:scheme}) requires the communication of the model $w_{t}$ and the computed gradients 
${\bf\Delta}w_t$ 
between all nodes in every iteration $t$. Since $w$ has to be synchronous in all nodes and ${\bf\Delta}w_{t+1}$ can
not be computed before $w_{t}$ is available, the entire communication has to be completed before the next 
iteration $t+1$. Naturally, one would try to overlap this communication (which can be done layer by layer)
with the compute times. However, there are several pitfalls to this strategy: (I) $w$ and ${\bf\Delta}w$ 
have the size of all weights in the neural network, which can be hundreds of megabyte\footnote{See table 
\ref{tab:bench} for details.}, 
(II) the compute times per iteration\footnote{See table \ref{tab:compute}.} 
are rather low and are decreasing further when scaling to more nodes 
(see section \ref{sec:matmul}), (III) communication can not start before the forward pass of the network 
has succeeded (practically cutting the overlay times by half). Ironically, faster compute units (e.g. newer
GPUs) in the compute nodes increase the fundamental problem, that the communication time exceeds the
compute time, after scaling to only a few nodes. Leaving valuable compute units idle. 
\begin{figure}
\includegraphics[width=\linewidth]{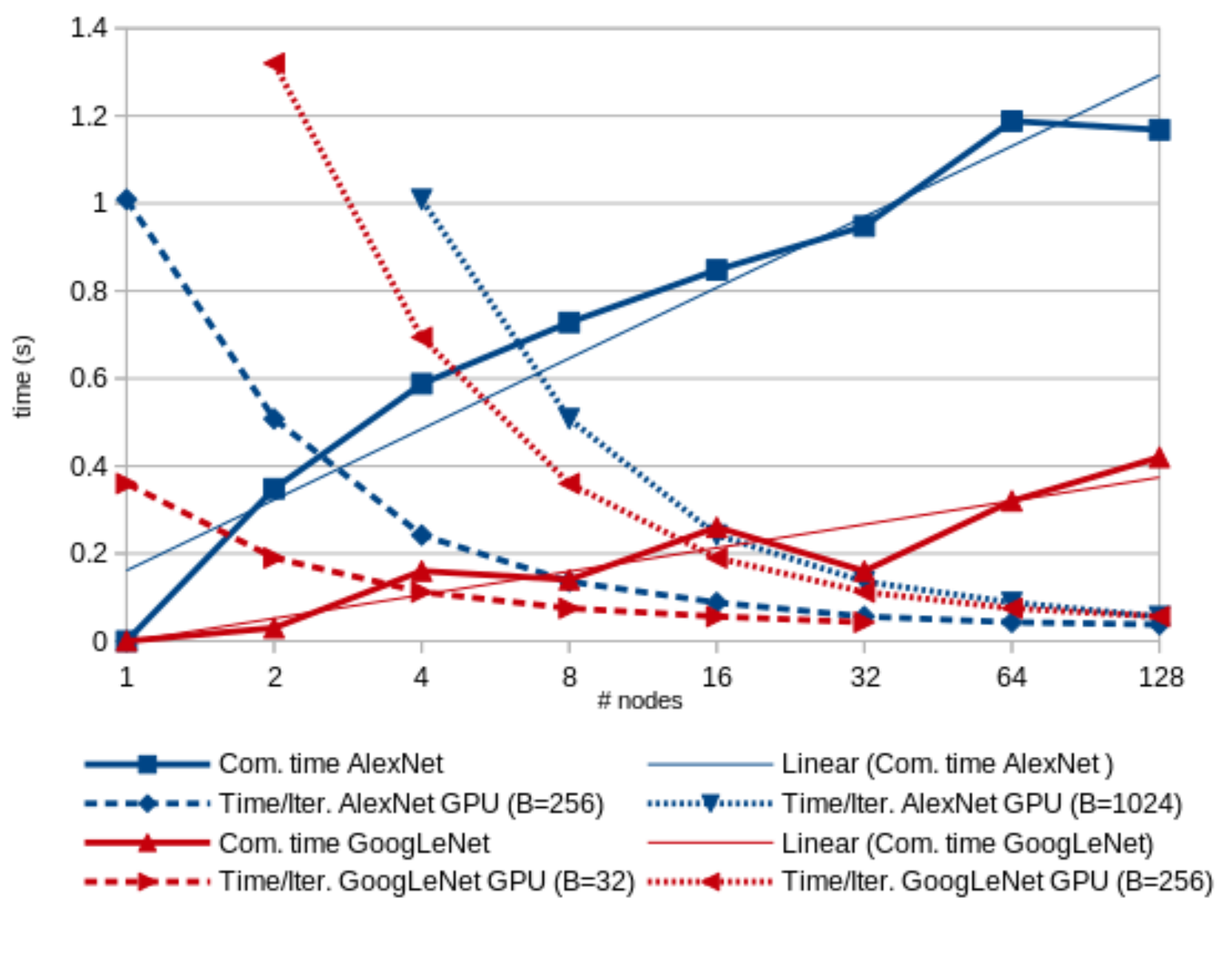}
\caption{Communication overhead for different models and batch sizes. The scalability stalls when the compute
times drop below the communication times, leaving compute units idle. Hence becoming an communication bound problem.
Results were generated using a binary tree communication scheme \cite{DBLP:journals/corr/IandolaAMK15}.
 \label{fig:overhead}}
\end{figure}
Figure \ref{fig:overhead} shows the strong divergence of communication- and compute times. Depending on the
model size, the training problem becomes {\bf communication bound} after scaling to only 4 to 8 nodes. This directly
correlates to the general scaling results shown in figure \ref{fig:Baseline}. Figure \ref{fig:overhead} also
shows, that the network layout has a large impact on the crucial communication/compute ratio: shallow networks
with many neurons per layer (like {\it AlexNet}) scale worse than deep networks with less neurons 
(like {\it GoogLeNet}) where
longer compute times meet smaller model sizes.  

\subsection{Limited Network Bandwidth}
\noindent Limited network bandwidth is one of the key bottlenecks towards the scalability of distributed 
DNN training.
Recently, there have been several approaches proposed to overcome this problem: 
e.g. \cite{DBLP:journals/corr/IandolaAMK15} 
introduced a binary communication tree, which reduces the network load to a maximum of $log_2(n)$ 
peer to peer model/gradient 
sends at a time. However, expecting linear speedups at the compute side, figure \ref{fig:overhead} shows that 
this approach will only move the 
intersection of the communication/compute ratio by a small factor, as the additional overhead is increasing with
the depth of the communication tree.\\     
\noindent Other methods try to reduce the model size before the communication. This can be done by (I) a
redesign of the network \cite{iandola2016squeezenet} - eliminating unused weights, 
(II) limiting the numerical precision of the model weights 
(\cite{gupta2015deep} has shown that one byte per weight is enough), (III) compression (which is available in 
\cite{intelcaffe}) or (IV) transmitting only sparse gradient and model information \cite{spring2016scalable}. 
All these methods have practical impact, moving the scalability by the factor of the model reduction rate. But non
of these approaches is able to solve the problem in principle. 
As model sizes are increasing much faster than the available 
network bandwidth, the communication overhead remains an unsolved problem.     


\section{Computational Costs and Scaling of Matrix Multiplications}
 \label{sec:matmul}
\noindent The previously discussed communication overhead is actually a well known problem that has recently been 
drawing more and more attention in the deep learning community 
\cite{DBLP:journals/corr/IandolaAMK15, iandola2016squeezenet,gupta2015deep,spring2016scalable}. But communication 
overhead is not the only problem preventing DNN scalability: 
there is an even more severe limitation, which turns out to be a hard theoretical constraint.     
\begin{figure}[h]
\includegraphics[width=\linewidth]{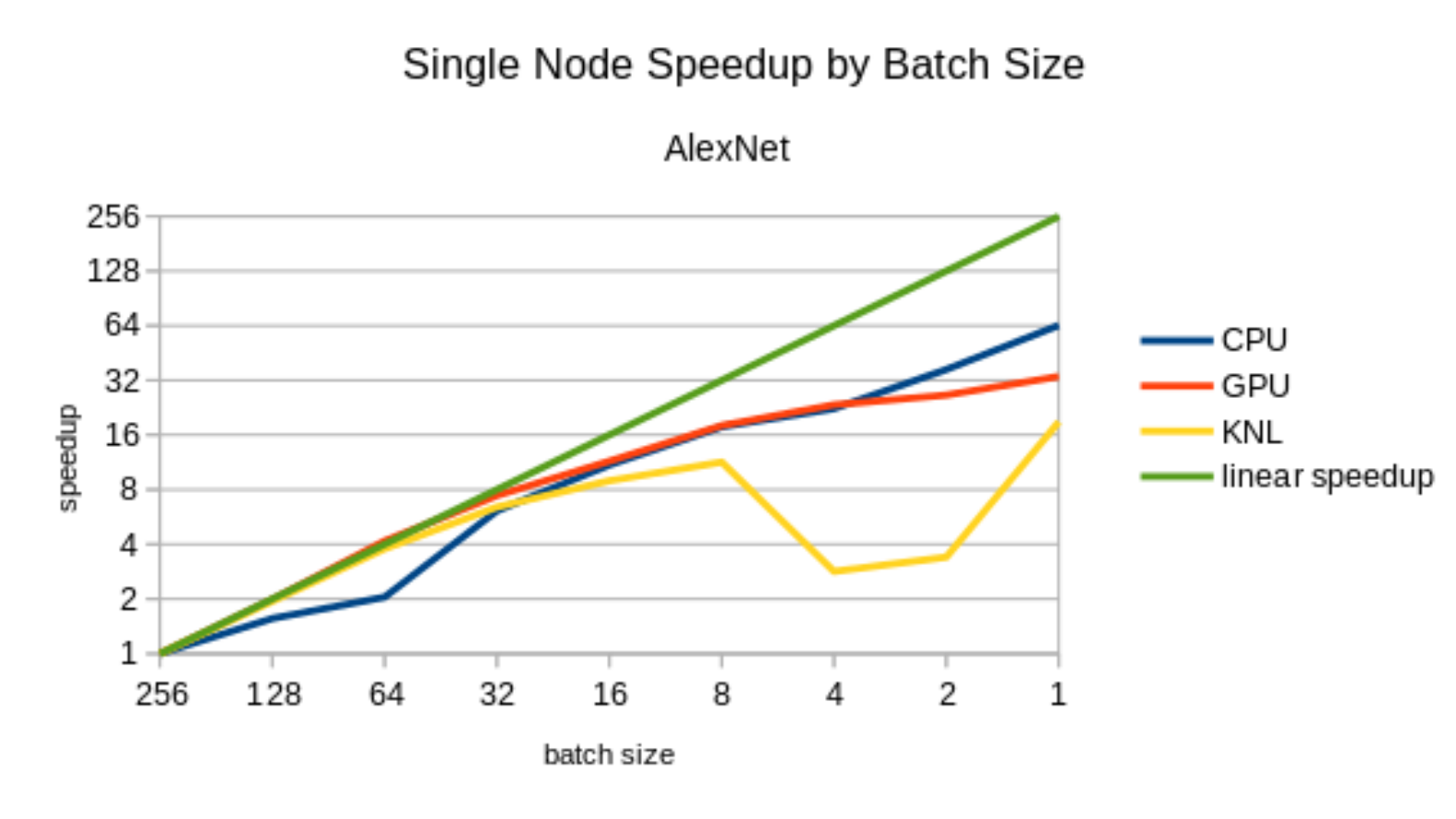}
\caption{Evaluation of the scalability assuming free communication (simulated by measuring the compute times at a single node at decreasing batch sizes). Results for different compute units. \label{fig:matrix_single_node}}
\end{figure}
We illustrate this problem by means of a simple experiment: assuming that the communication in the distributed DNN
training was free, one would expect close to linear strong scaling\footnote{because distributed SGD is 
data-parallel} properties. However, figure \ref{fig:matrix_single_node} shows, that this is not the case.
Again, scalability stalls after only a few nodes. While it is obvious that it is not possible to 
split the global batch into local batches $b<1$, thus imposing a {\bf strict scalability limit} at $n=B$,
the limitation induced by the batch size are taking effect even for $b>>1$.  
To allow further investigation of these results, we provide a 
layer by layer analysis on the computational complexity and scalability of out benchmark networks.     


\subsection{Layer by Layer Analysis}
\label{sec:layerbylayer}
\begin{figure}[ht]
\includegraphics[width=\linewidth]{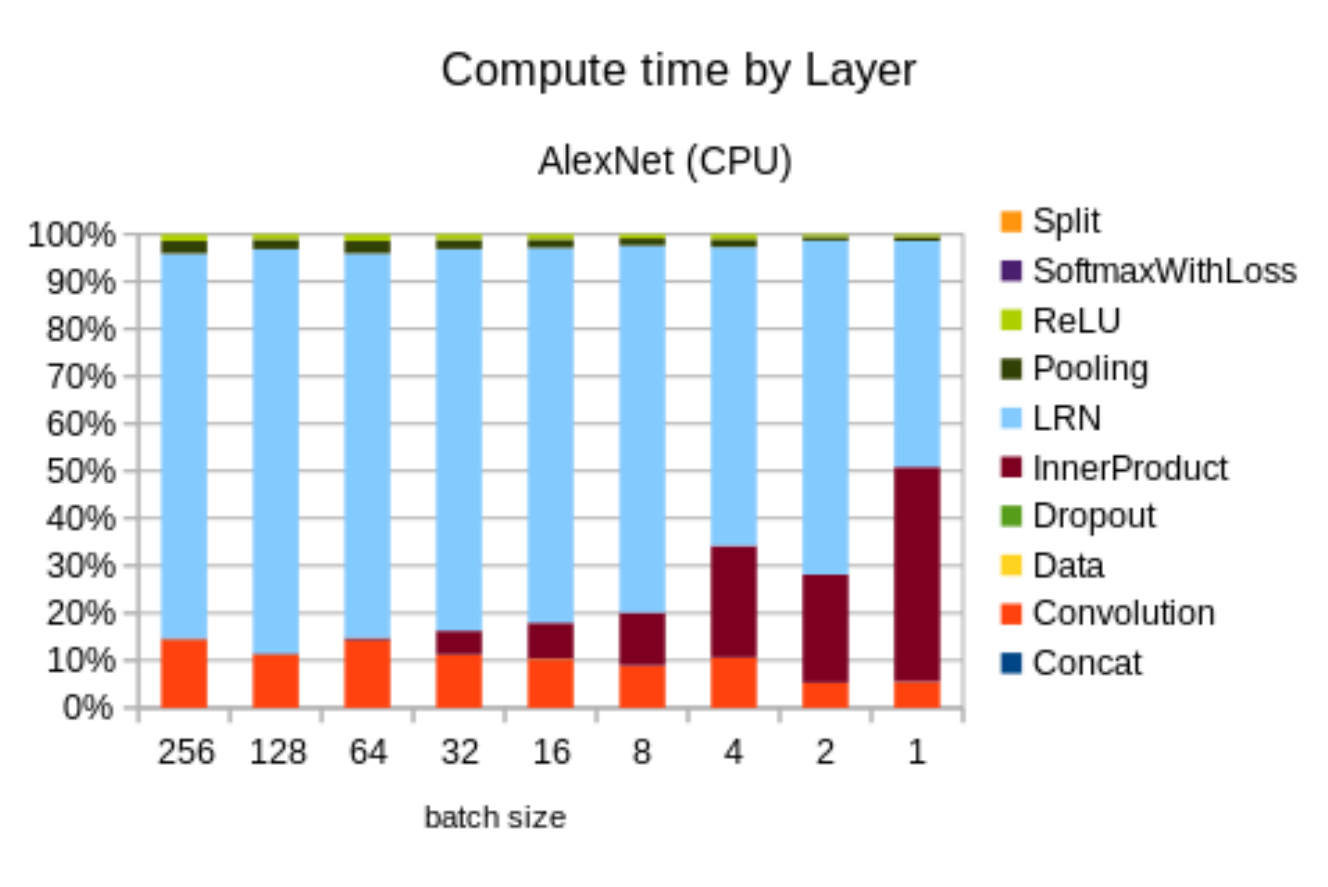}
\includegraphics[width=\linewidth]{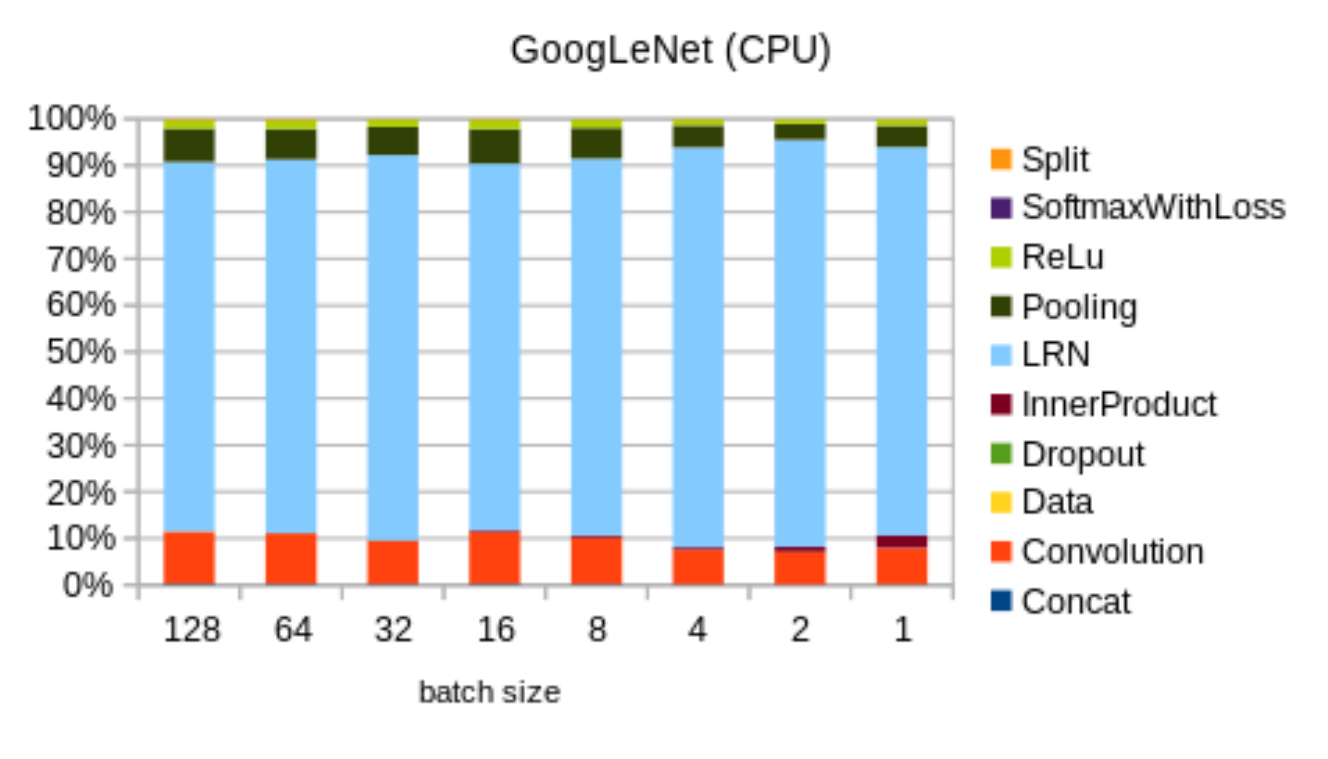}
\caption{Evaluation of the relative compute time for each layer type (several layers of the same type are accumulated) per training iteration on a single node CPU based system. Top: results for {\it AlexNet}. 
Bottom: results for {\it GoogLeNet}. \label{fig:layer_time_cpu_alex}}
\end{figure}
\begin{figure}[h!]
\includegraphics[width=\linewidth]{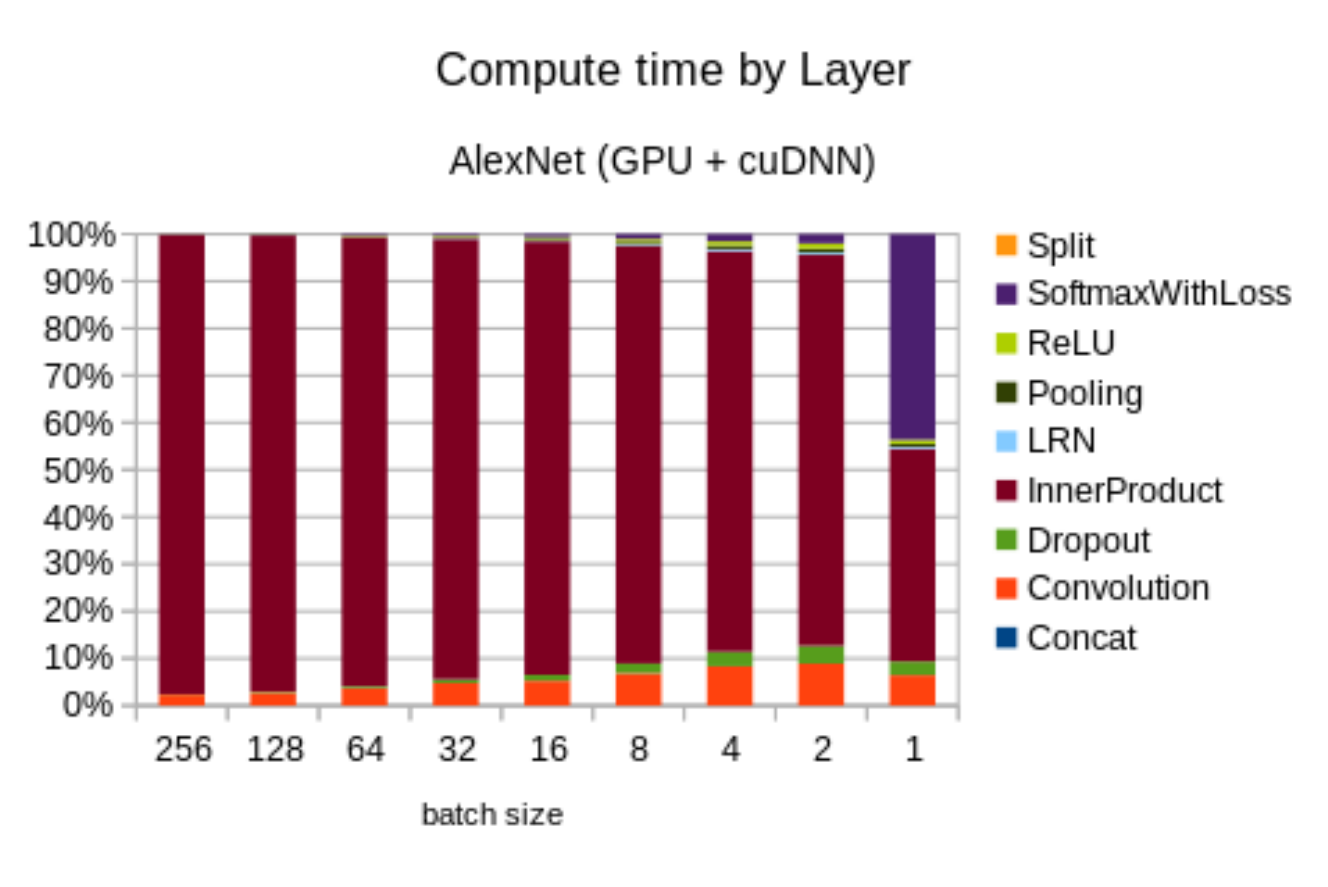}
\includegraphics[width=\linewidth]{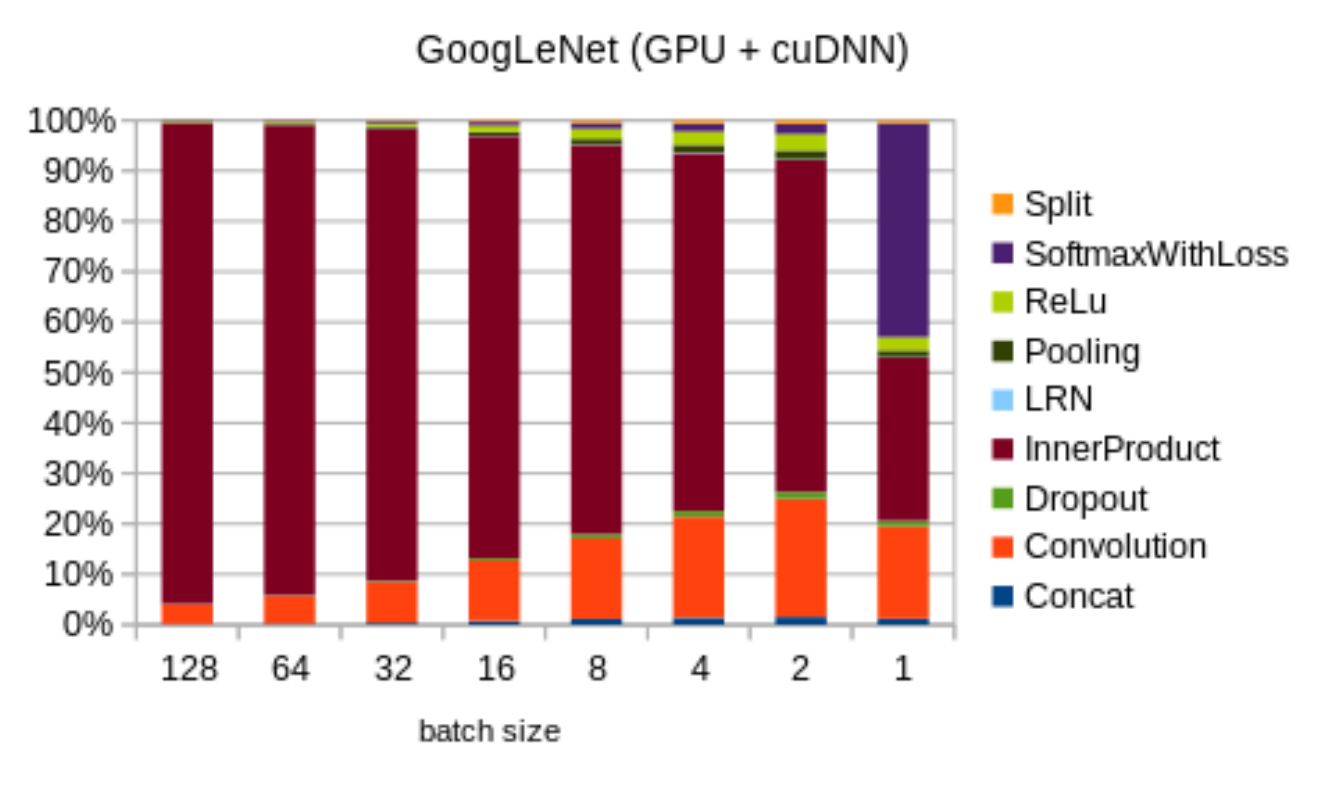}
\caption{Evaluation of the relative compute time for each layer type (several layers of the same type are accumulated) per training iteration on a single node GPU based system (one K80). Top: results for {\it AlexNet}. Bottom: results for {\it GoogLeNet}. \label{fig:layer_time_gpu_alex}}
\end{figure}
\begin{figure}[ht!]
\includegraphics[width=0.9\linewidth]{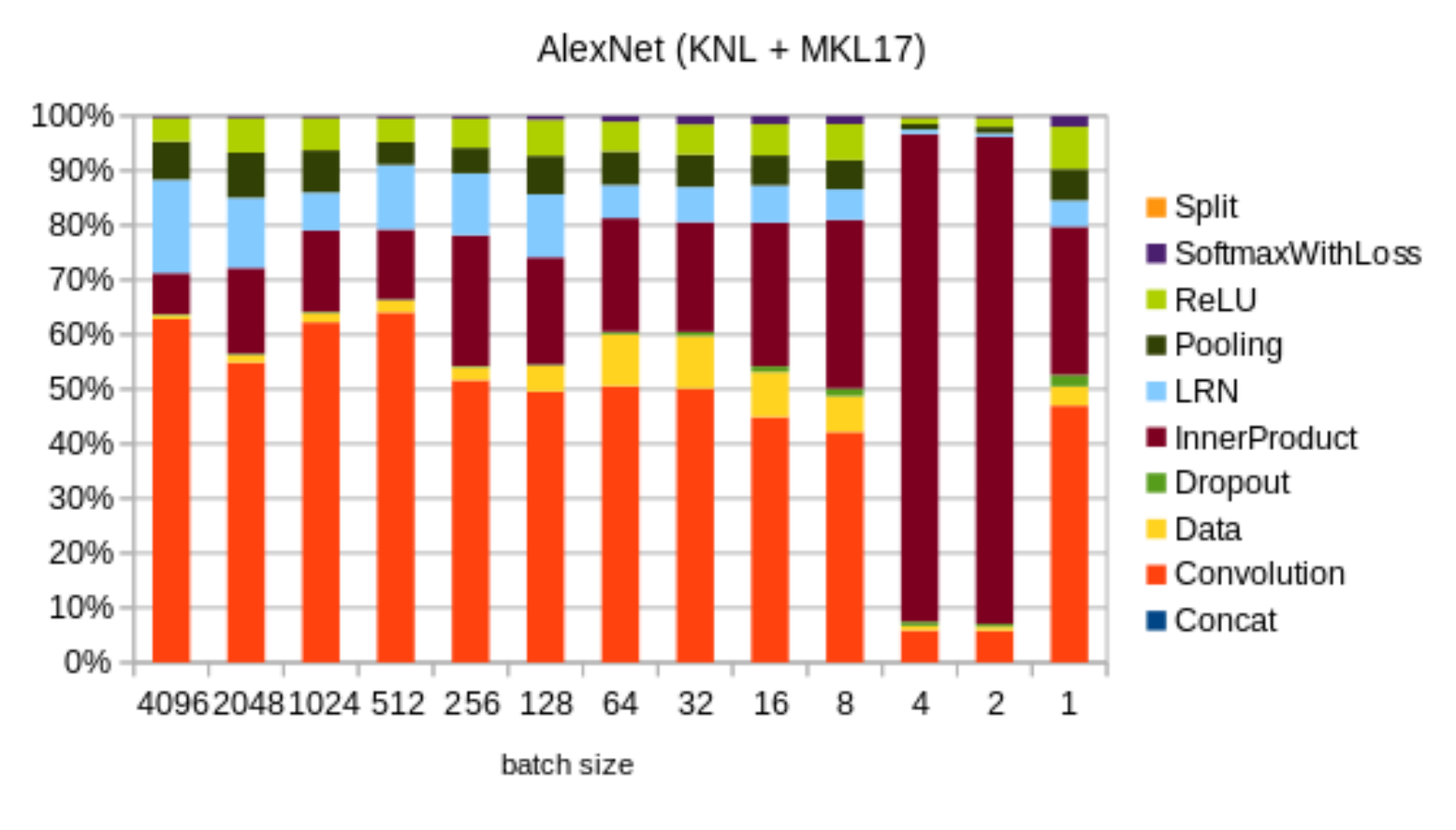}
\includegraphics[width=0.9\linewidth]{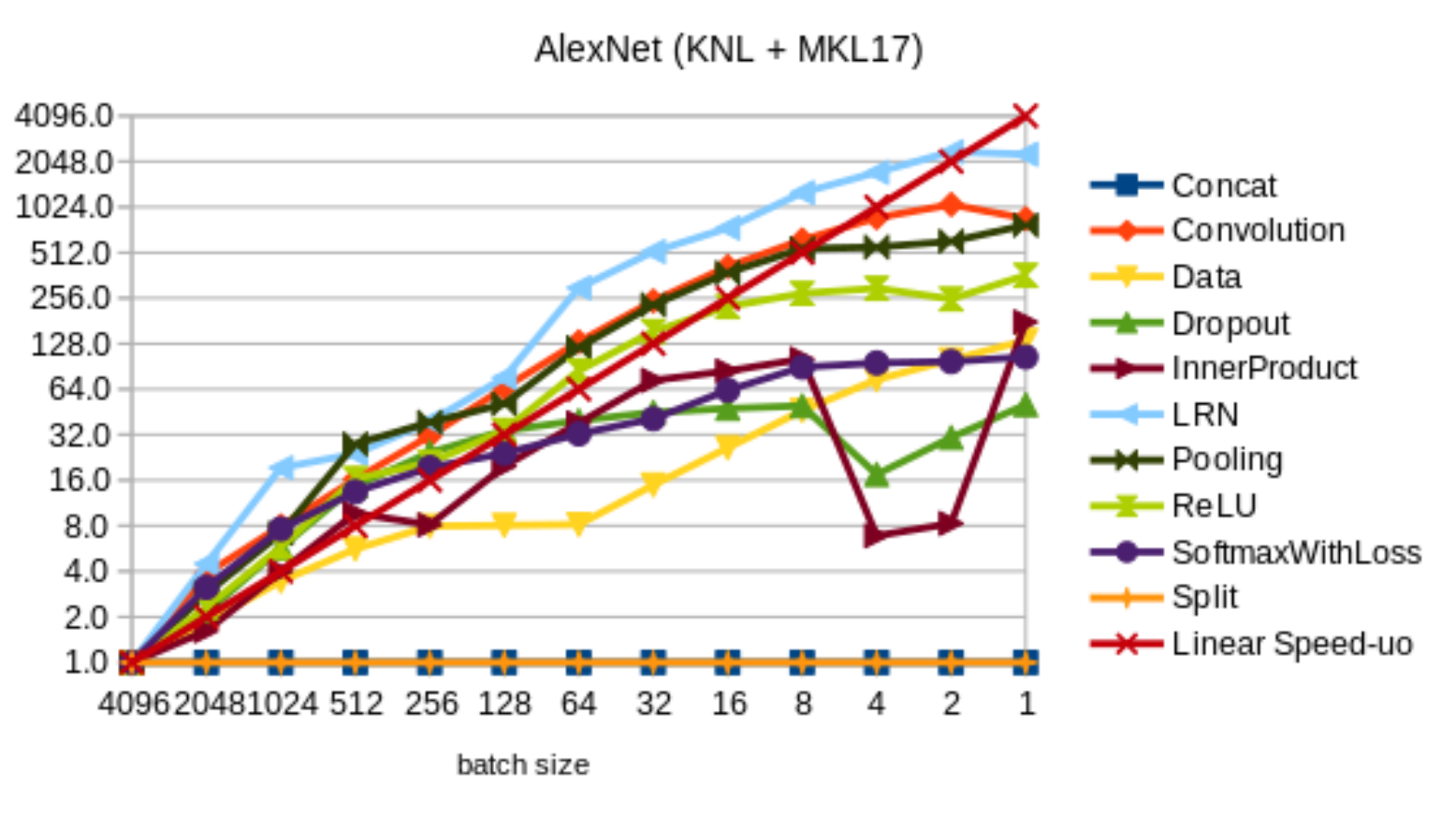}
\caption{Additional layer by layer results for the KNL (see section \ref{sec:layerbylayer} for details). Top: Proportional compute times by layer type and batch size for {\it AlexNet}. Bottom: Speedups by layer type ans batch size for {\it AlexNet}.  \label{fig:knl}}
\end{figure}
\begin{figure}[ht!]
\includegraphics[width=0.8\linewidth]{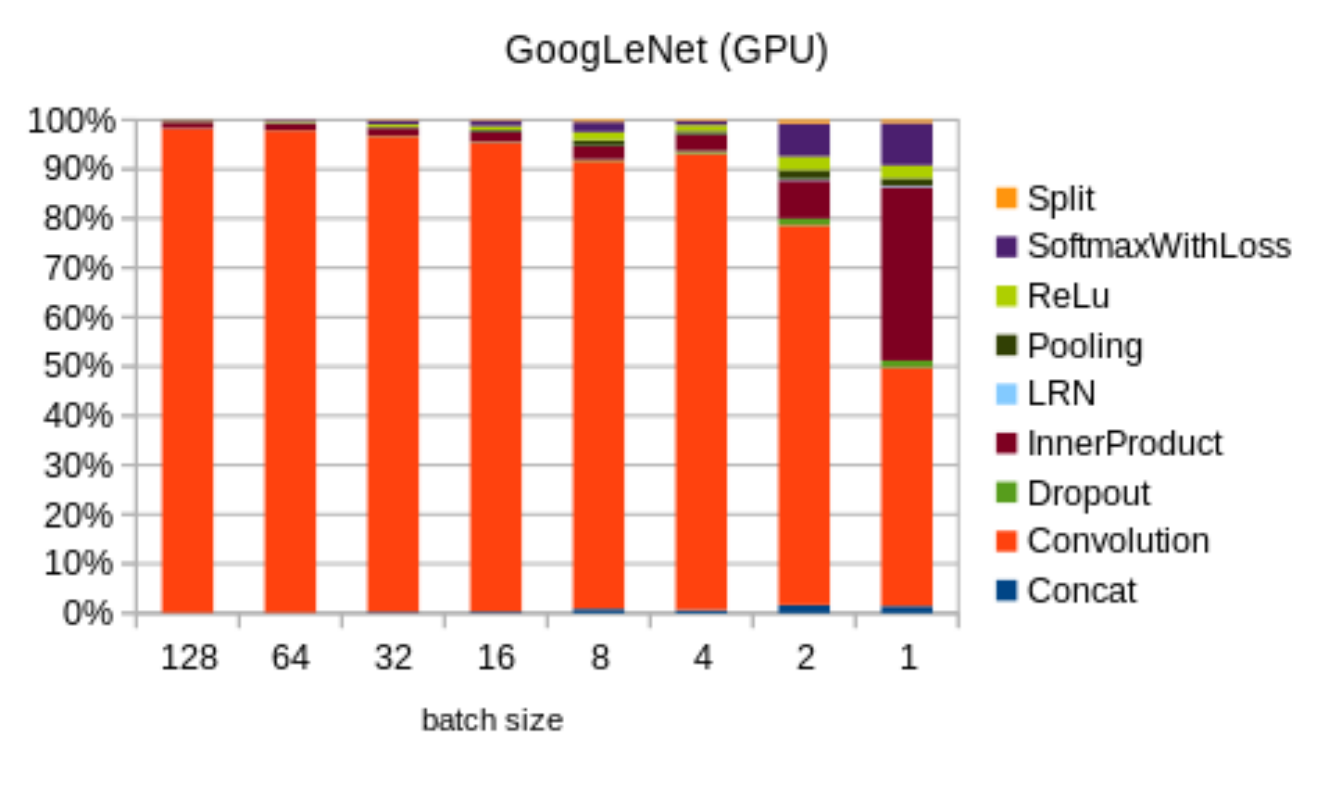}
\caption{Layer by Layer analysis for {\it GoogLeNet} without cuDNN. \label{fig:layer_time_gpu_nocuDNN}}
\end{figure}

\noindent Figure \ref{fig:layer_time_cpu_alex} shows the analysis for DNN training on CPUs. The dominance of the
Local Response Normalization layer (LRN) is caused by a rather poor multi-threaded 
implementation\footnote{This problem has been fixed by the MKL17 implementation as shown figure \ref{fig:knl}.} 
in {\it Caffe}
and is neglectable in terms of scalability (as shown in figure \ref{fig:layer_time_cpu_alex}). More interesting
is the growing portion of compute time spend in the {\it InnerProduct} (= Fully-Connected)\footnote{The changing name convention is due to the naming used in Caffe.}  layer.
Figure \ref{fig:layer_time_gpu_alex} shows the same tendencies for layer computations on GPUs, where the LRN layer
has no significant impact.
\noindent Yet another interesting observation can be made in figure \ref{fig:layer_time_gpu_nocuDNN}, 
which shows the impact
of the convolution optimization of the cuDNN\footnote{The optimization strategy is also available in MKL17, as shown in figure \ref{fig:knl}.} library used in figure \ref{fig:layer_time_gpu_alex}.
\noindent Even more evident than the relative compute portions of the different layer types shown in figures
\ref{fig:layer_time_cpu_alex}, \ref{fig:layer_time_gpu_alex}, \ref{fig:layer_time_gpu_nocuDNN} and \ref{fig:knl},
are the scaling properties by the different layer types. 
Figure \ref{fig:layer_time_cpu_alex} depicts these for DNN training on 
CPUs (see figure \ref{fig:knl} for results on the new Xeon-Phi): all but one layer types show almost perfect linear
scaling. Only the significantly compute intensive {\it InnerProduct} layer scales poorly for batch sizes $b<64$, 
which is equivalent to scaling to only $n>4$ nodes for the original
batch size $B=256$. On the GPU, the crucial {\it InnerProduct}
layer scales much better than on the CPU, but
still fails linear speedup as we can see accelerations factors around $32$x at $256$ nodes.
Again, the speedup stalls for batch sizes $b\leq32$.

\begin{figure}[ht]
\centering
\includegraphics[width=\linewidth]{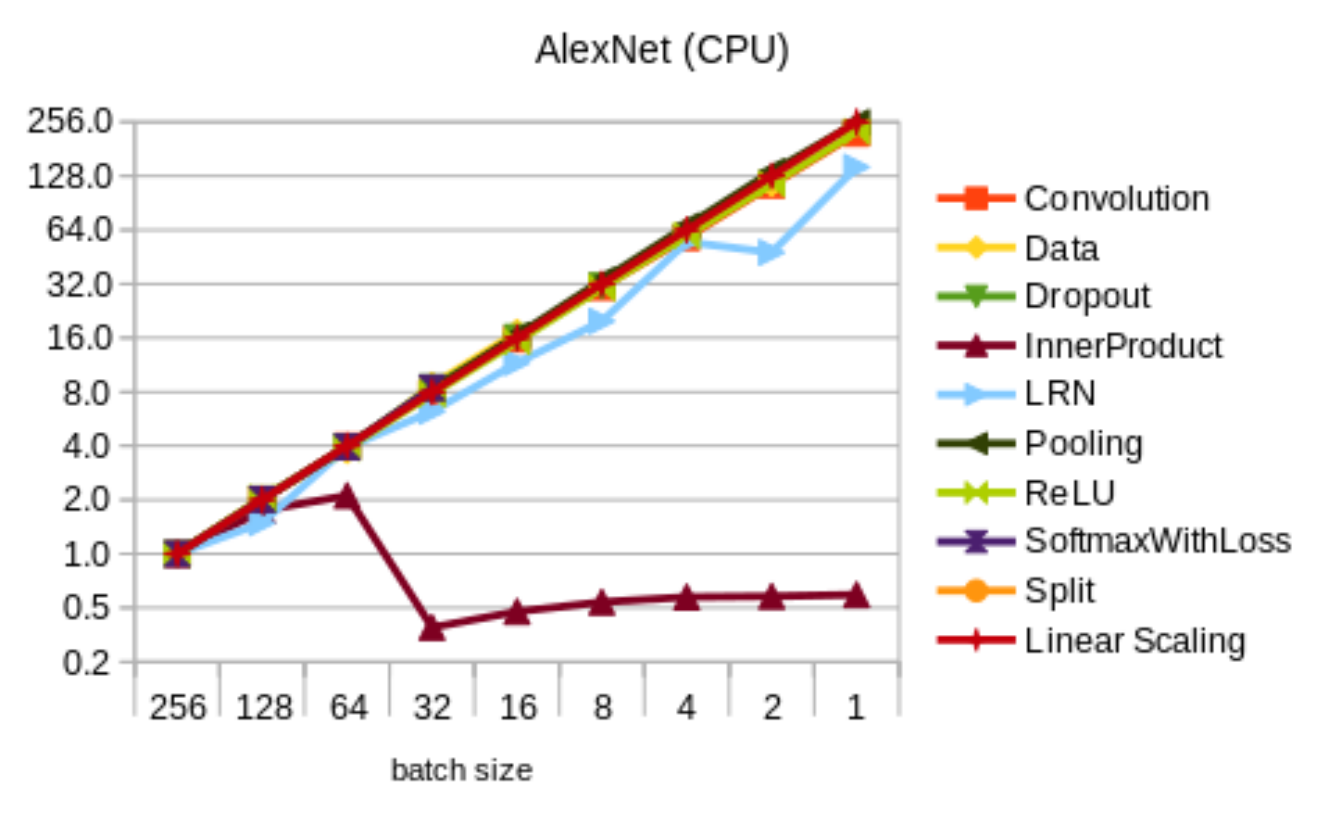}
\includegraphics[width=\linewidth]{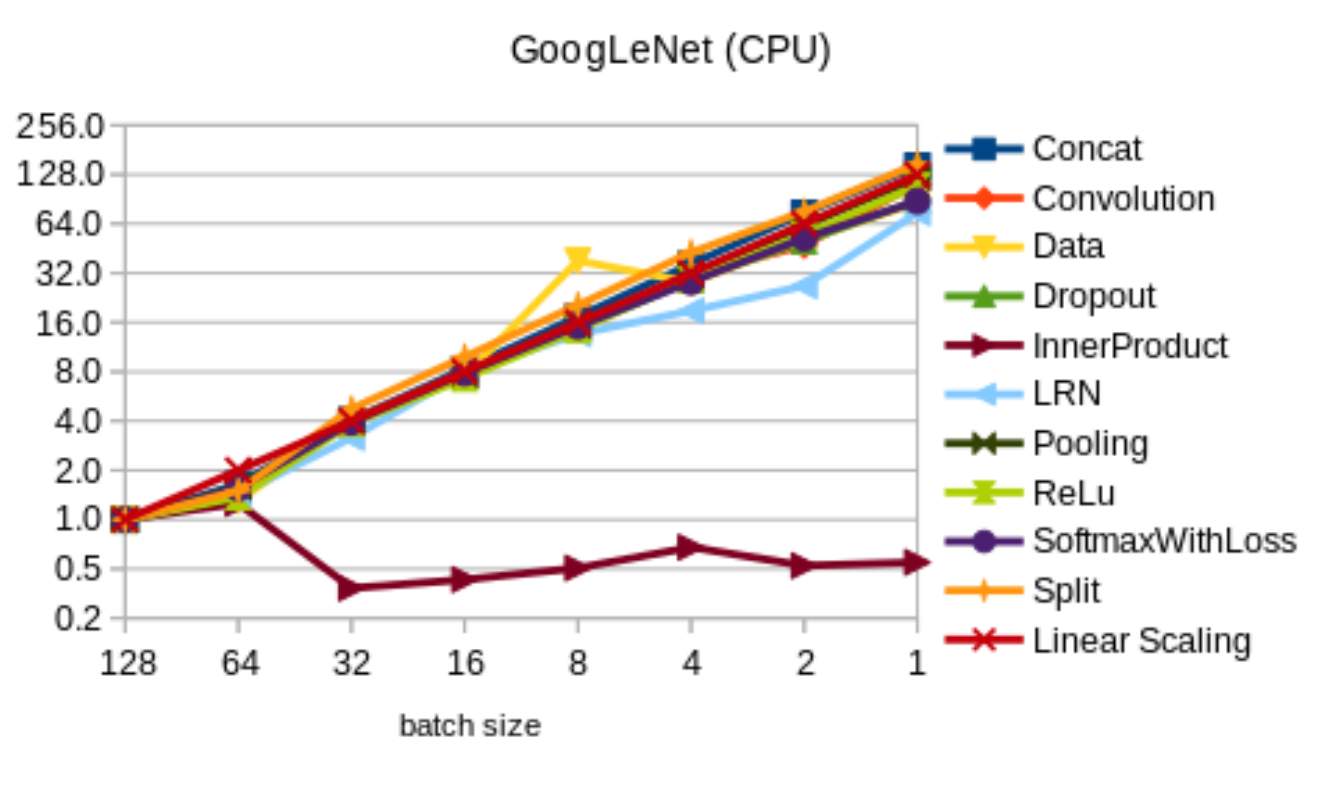}

\caption{Speedup achieved by reducing the batch size - computed from the results in Fig. \ref{fig:layer_time_cpu_alex}. Top: results for {\it AlexNet}. Bottom: results for {\it GoogLeNet}. }
\end{figure}

\begin{figure}[ht]
\includegraphics[width=\linewidth]{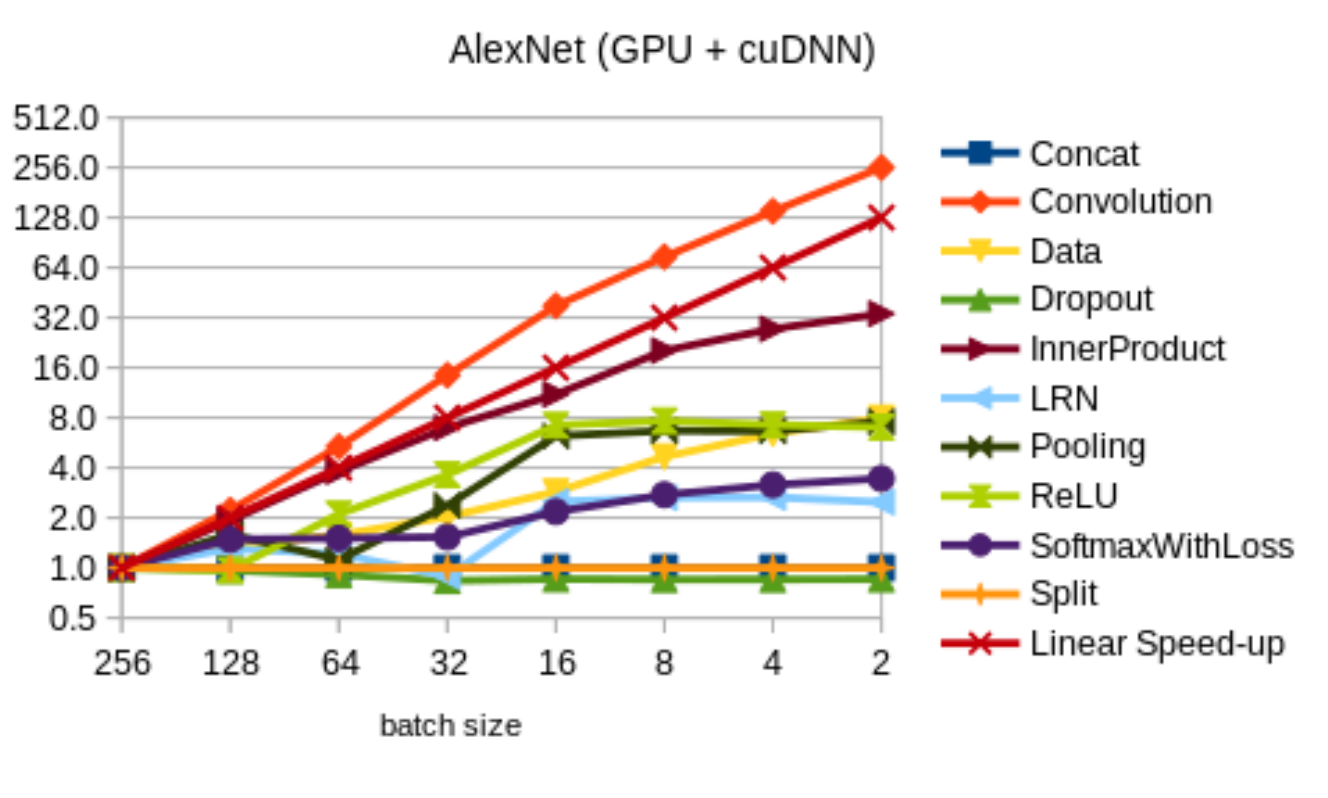}
\includegraphics[width=\linewidth]{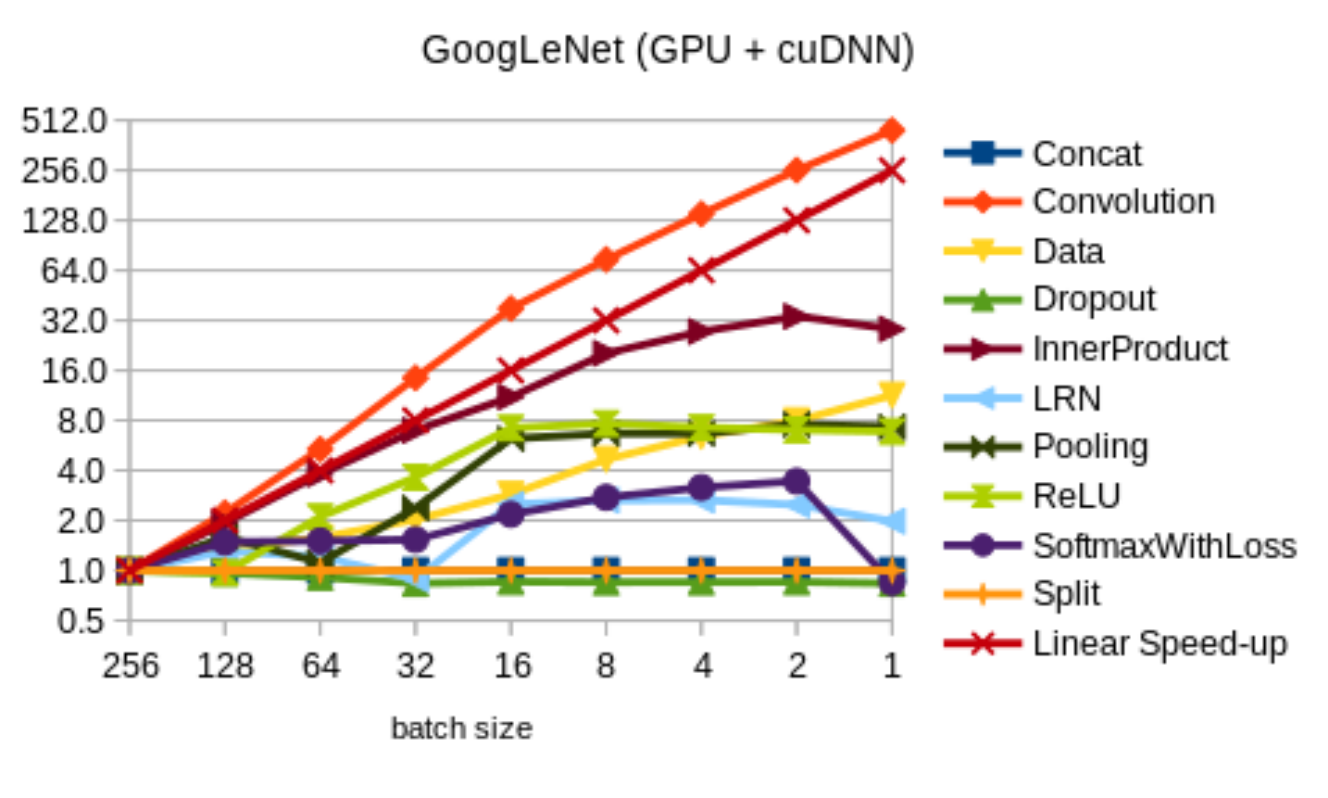}
\caption{Speedup achieved by reducing the batch size - computed from the results in Fig. \ref{fig:layer_time_gpu_alex}. Top: results for {\it AlexNet}. Bottom: results for {\it GoogLeNet}. \label{fig:speedup_gpu_alex}}
\end{figure}

\subsection{Scaling Fully-Connected Layers}
\begin{figure}
\includegraphics[width=\linewidth]{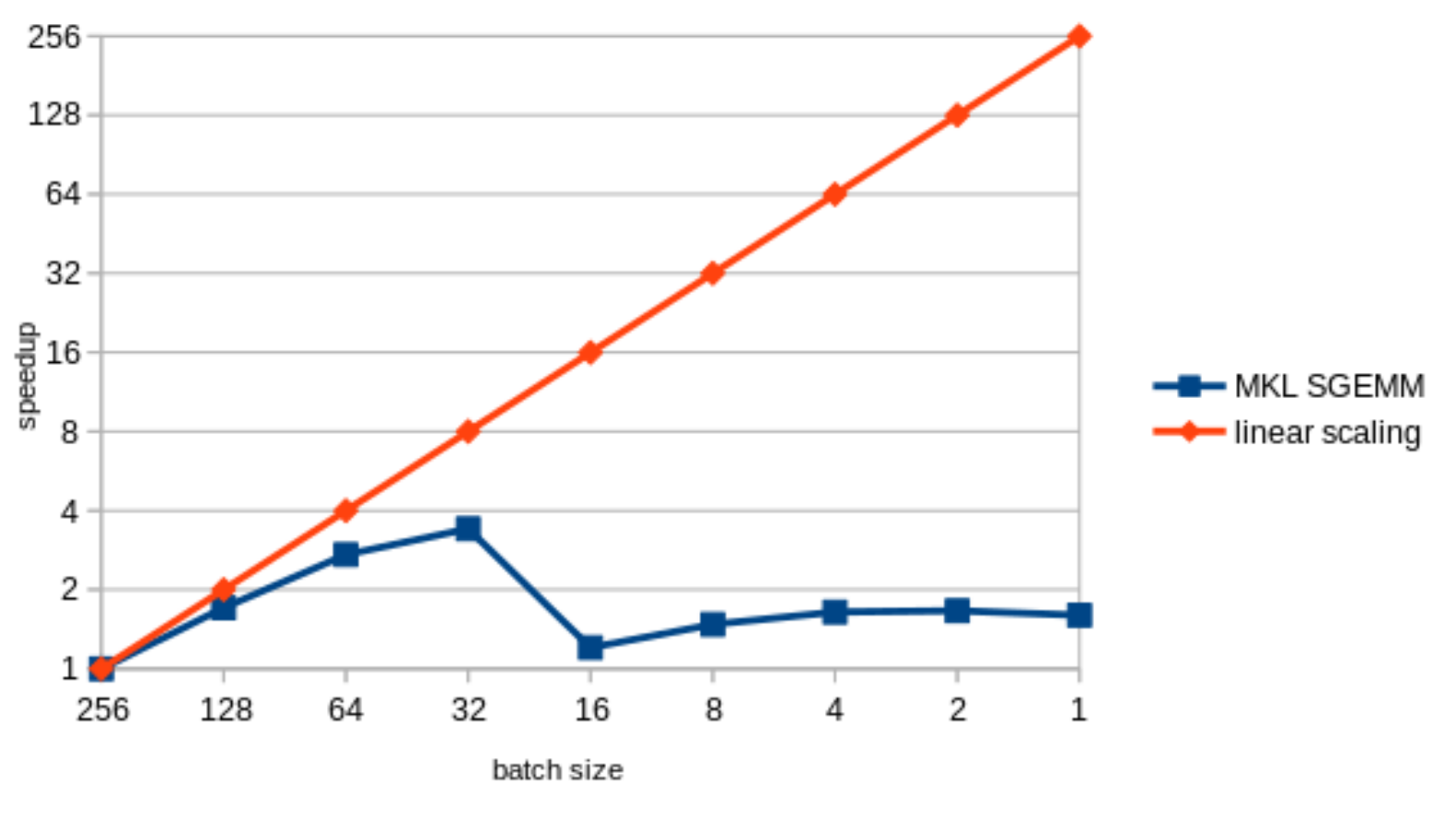}
\caption{MKL SGEMM: impact of the batch size $b$ on the MKL sgemm speedup performance for matrix multiplications 
with the shape $b\times 4096 * 4096\times 9192$. These matrix shapes correspond to the sgemms computed in the largest
{\it Fully-Connected} layer of {\it AlexNet}. \label{fig:sgemm}}
\end{figure}

\begin{table}[h!]
\centering
\begin{tabular}{l|cc}
Layer & \# operations & matrix sizes \\
\hline
Fully Connected & $1$ & $b\times I * I\times O$  \\
Convolutional  & $b$&  $C\times I * I \times Z$  \\
Softmax & $b$&  $I\times 1 * 1\times 1$\\
& &
\end{tabular}
\begin{tabular}{ll}
&\textbf{Definitions:}\\
I: & Input size from top layer\\
O: & Output size of this layer\\
b: & local Batch size (train or validation)\\
C: & Number of filters \\
c: & Number of input channels (RBG image: $c=3$)\\
P: & Patch size (i.e. pixel)\\
k: & kernel size\\
Z: & Effective size after kernel application.\\
& For convolution $Z:=\left(\sqrt{P}-\lfloor(k/2)\right)^2$\\
\end{tabular}
\caption{Size and number of of the matrix multiplications (sgemm) per forward pass for selected layers.
\label{tab:matsize}}
\end{table}
\noindent The layer by layer analysis revealed the impact of the {\it Fully-Connected} (FC) layers on the overall 
scalability.
FC layers are the ``conventional'' neural layers in the deep network architecture, where
the actual decision boundaries of a classification problem are modeled. Typically, FC layers hold a large number 
of neurons which are connected to all inputs. Computationally, FC layers perform a single\footnote{Actually there is a second small matrix multiplication for the computation of the bias which we neglect here.}  matrix 
multiplication per pass.
\noindent Table \ref{tab:matsize} shows the impact of the batch size $b$ on the size, shape and number of matrix
multiplications. While $b$ only affects the number of matrix operations for {\it Convolutional} layers (which can 
implemented task-parallel \cite{krizhevsky2014one}), it directly reshapes the left-hand matrix in the FC sgemm
operation in a very unfavorable way. For typically large $I$ and $O$ (e.g. for a layer in {\it AlexNet} we find 
$I=4096, O=9192$), $b=B/n$ decreases from $B=256$ - producing ``degenerated'' (maximal non-square) matrices.
This ``degeneration'' hurts the {\it Inner Parallelization} of the matrix multiplication (see 
section \ref{sec:PSGD}), where
the sgemm is either multi-threaded by the MKL Blas Library or parallelized via cudaBlas. Both implementations
have an optimal performance for square matrices and suffer from the ``degeneration'' \cite{demmel2013communication}.
Hence, speedups gained by the {\it Outer Parallelization} (the data-parallel SGD) start to harm the performance 
of {\it Inner Parallelizations} and cause a ``scalability deadlock''. 
Figure \ref{fig:sgemm} shows the impact of $b$ on the MKL sgemm speedup performance. It is not
surprising, that this evaluation shows exactly the same speedup characteristic as the overall communication free
scaling experiment in fig. \ref{fig:matrix_single_node}.             

\subsection{Increasing the global Batch Size}
\begin{figure}[ht]
\includegraphics[width=\linewidth]{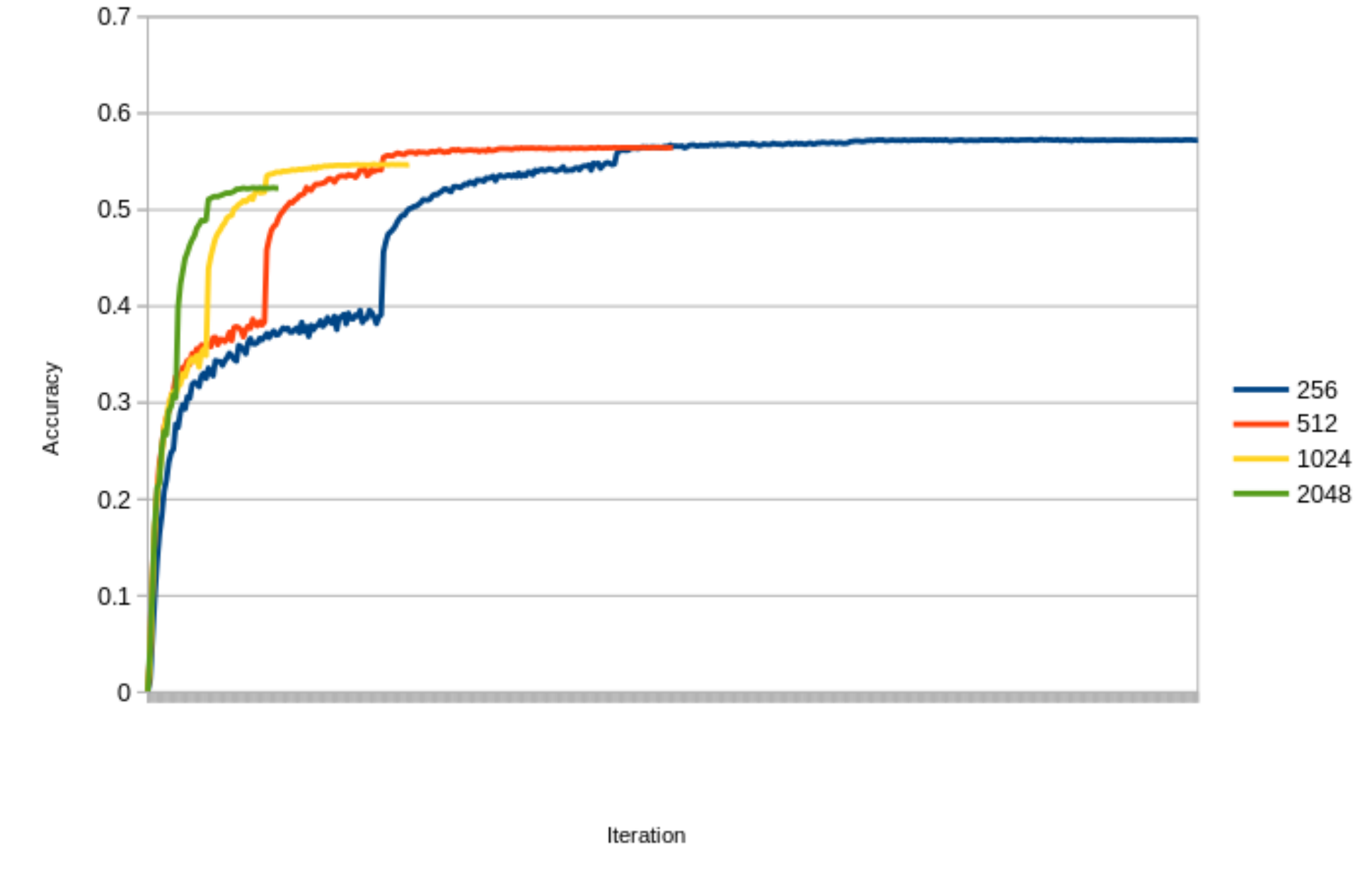}
\caption{Full validation accuracy plot for {\it AlexNet} with different large batch sizes. Settings 
[$B=256, \epsilon=0.01, iter=450k$], [$B=512, \epsilon=0.02, iter=225k$], [$B=1024, \epsilon=0.04, iter=112k$], 
[$B=2048, \epsilon=0.08, iter=56k$]    \label{fig:batch_acc}}
\end{figure}

\noindent A simple way to overcome the stalling scalability beyond 8 nodes (or $b<32$), has recently been suggested
in \cite{DBLP:journals/corr/IandolaAMK15} and is also utilized in \cite{intelcaffe}: increasing the global batch
size $B$ to the extend, that the worker batch size keeps an effective size 
$b\geq32$ for the {\it Inner Parallelization}.  

\begin{figure}[ht]
\includegraphics[width=\linewidth]{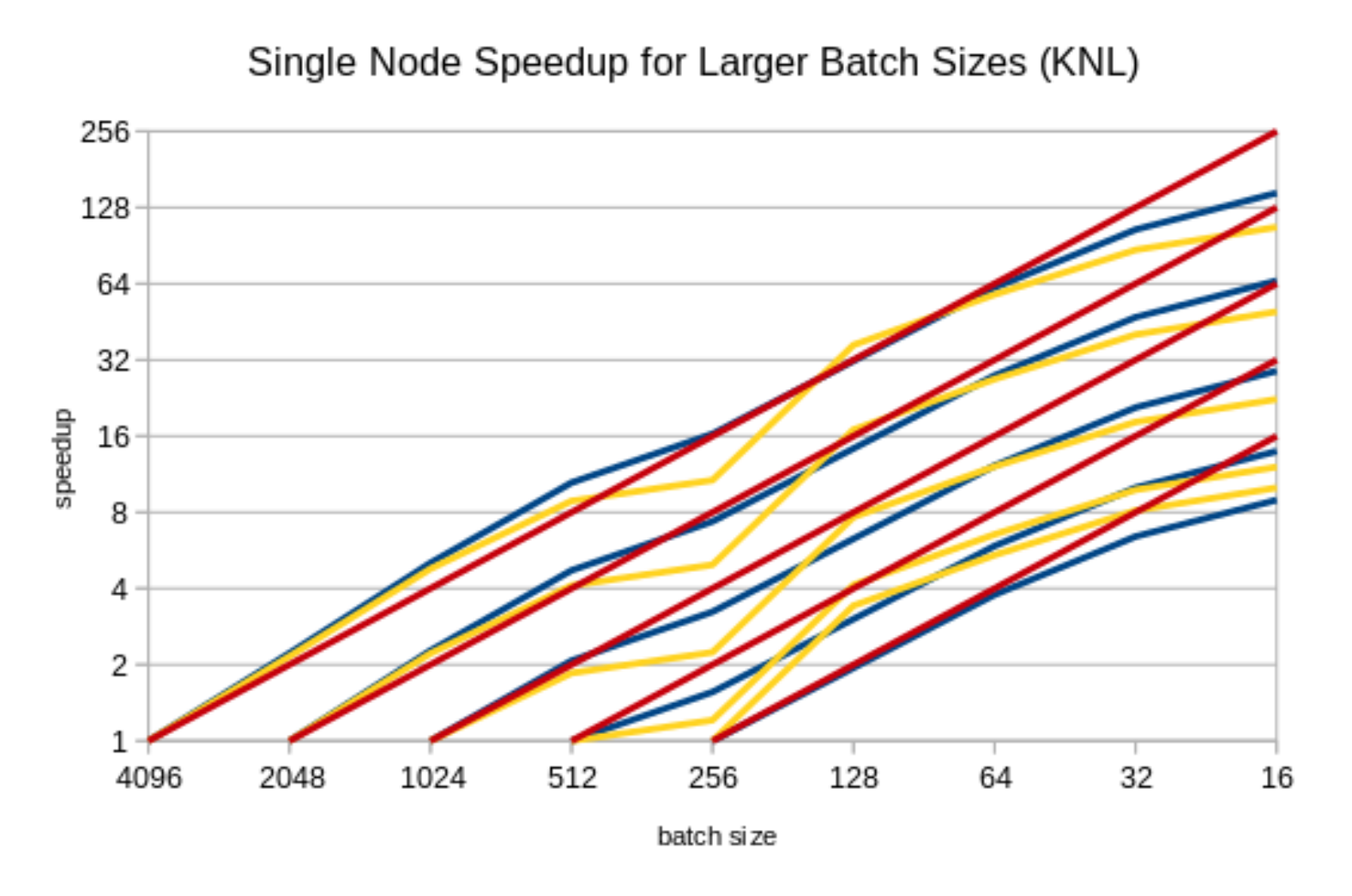}
\caption{Scaling properties for increased global batch sizes in the ``free communication'' scenario. Yellow lines
show the results for {\it AlexNet}, blue lines {\it GoogLeNet} and red lines indicate perfect linear speedup. NOTE: All 
speedups are computed with respect to the compute time of the enlarged global batch sizes, not the original batch
sizes ($B=256$ for {\it AlexNet} and $B=32$ for {\it GoogLeNet}). \label{fig:largeB}}
\end{figure}
\noindent Figure \ref{fig:largeB} shows, that this strategy is actually providing almost perfect linear speedup 
up to 128 nodes for a global batch size of $4096$. However, these results have to be taken with strong caution:
Increasing the global batch size also increases the computational complexity of the problem linearly.
Beyond a certain batch size, SGD will not converge significantly faster in terms of the number of iterations. 
Hence, large batch sizes will increase the computation time per iteration while the number of iterations stays 
constant. In order to reduce the number of iterations till convergence, one would have to increase the step size
as well.
The authors of \cite{DBLP:journals/corr/IandolaAMK15} argue, that larger batch sizes will provide more stable 
gradient information which should allow larger step sizes. If it was possible to increase the step size in 
the same way this is done with the batch size, one would yield perfect linear scaling.\\
\noindent Sadly, this is hardly the case. Figure \ref{fig:batch_acc} shows the accuracy plots for {\it AlexNet}, 
computed till full convergence with differently large global batch sizes. 
The experiments were performed on a single 
 KNL node to avoid possible interferences in a distributed setting\footnote{The KNL provides enough
memory for such large batch sizes. On common GPUs with 12GB memory, the bach size limit is $b=256$ for 
{\it AlexNet} and $b=128$ for {\it GoogLeNet}.}. The step sizes $\epsilon$ were increased 
according to the batch size as suggested by \cite{DBLP:journals/corr/IandolaAMK15}, 
while the number of iterations has been decreased by the same factor. \\
The results are quite disappointing: while we reach linear speedup as expected, the validation accuracy 
is suffering significantly:
\begin{table}[h!]
\centering
\begin{tabular}{l|ccc}
batch size & speedup &  step-size &accuracy \\
\hline
$256$ & $1$ & $\epsilon=0.01$ & $57.2\%$ \\ 
$512$ & $2$ & $\epsilon=0.02$ & $56.4\%$ \\
$1024$ & $4$ & $\epsilon=0.04$ &$54.7\%$ \\
$2048$ & $8$ & $\epsilon=0.08$ &$52.2\%$ \\
\end{tabular}
\caption{Effect of choosing larger step-sizes $\epsilon$ on the resulting test/validation accuracy. Our experiments show, that larger batch sizes con not compensate for the loss in accuracy. }
\end{table}
These experimental results confirm the theoretic analysis by \cite{keskar2016large}, who showed that large batch sizes
lead to sharp minima with poorer generalization properties.  
Considering, that an early stopping of the original problem when reaching the according error rates yields almost 
the same speedup as the parallelized large batch variants, shows that this approach might not be suitable to
solve the scaling problem of the matrix multiplications.            

\subsection{Non-Scaling Layers}
\begin{figure}[ht!]
\includegraphics[width=\linewidth]{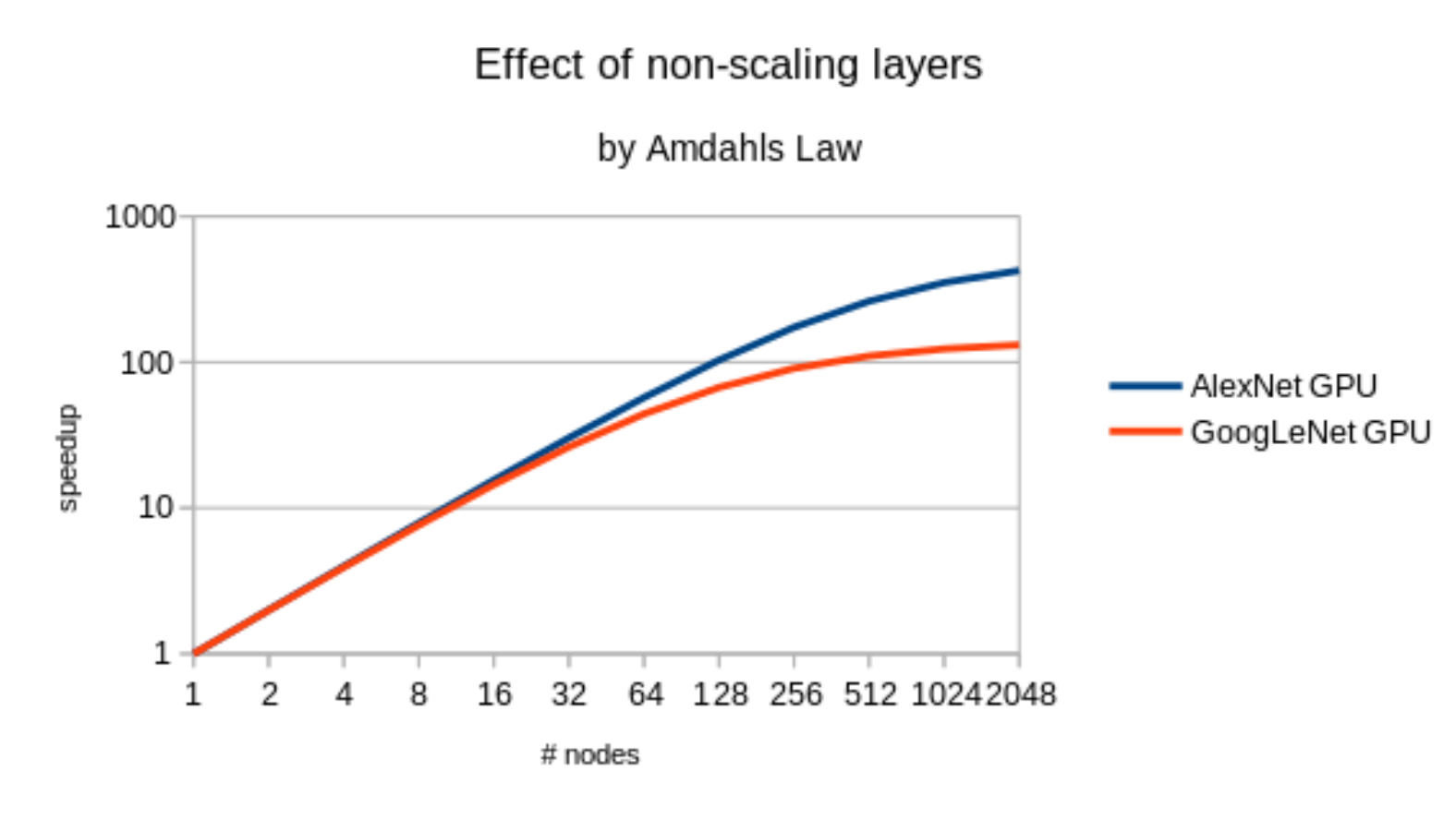}
\caption{Effect of non-scaling layers to the overall scalability after Amdahl's Law.  \label{fig:amdahl}}
\end{figure}

\noindent Figure  
\ref{fig:speedup_gpu_alex} also shows very poor scalability for some layers like {\it Dropout, Pooling} 
or {\it LRN}. This is mostly due to the fact that these layers are computed so fast, that the latency of loading the data to
the GPU becomes the dominant constant factor.
Overall, these particular layers consume only a marginal portion of the total compute 
time\footnote{See figure \ref{fig:layer_time_gpu_alex}.} (0.1\% for {\it AlexNet} and 0.3\% for {\it GoogLeNet}, 
assuming that all other layers are parallelizable). Applying {\it Amdahl's Law} shows in figure \ref{fig:amdahl}, 
that this still affects the scalability in the long run. Again, scalability begins to stall at $n>32$.  

\section{Parallel Training Data Access}
\label{sec:distdata}
\noindent So far, the analysis in the previous sections neglected another crucial bottleneck towards scalable 
distributed 
DNN training: the distribution of the training data (a.k.a. the batches) to the worker nodes. We specifically 
avoided this problem in all prior experiments by holding copies of the entire training set on local SSDs on
every worker node. However, this approach not only requires the availability of NVRAM (or other high speed local
storage) at every node, 
it is also very inefficient to copy hundreds of gigabyte\footnote{{\it ImageNet 1000} 
\cite{russakovsky2015imagenet} has $\sim$ 150GB of training data, larger real world problems easily exceed 
many terabytes.} to each worker node before the actual training can be started.        

\subsection{Network Bandwidth Revised} 
\noindent Using a centralized storage for the training data, like a database server or a distributed file system, 
would offer 
a more convenient and resource effective solution. 
\begin{figure}[ht!]
\includegraphics[width=\linewidth]{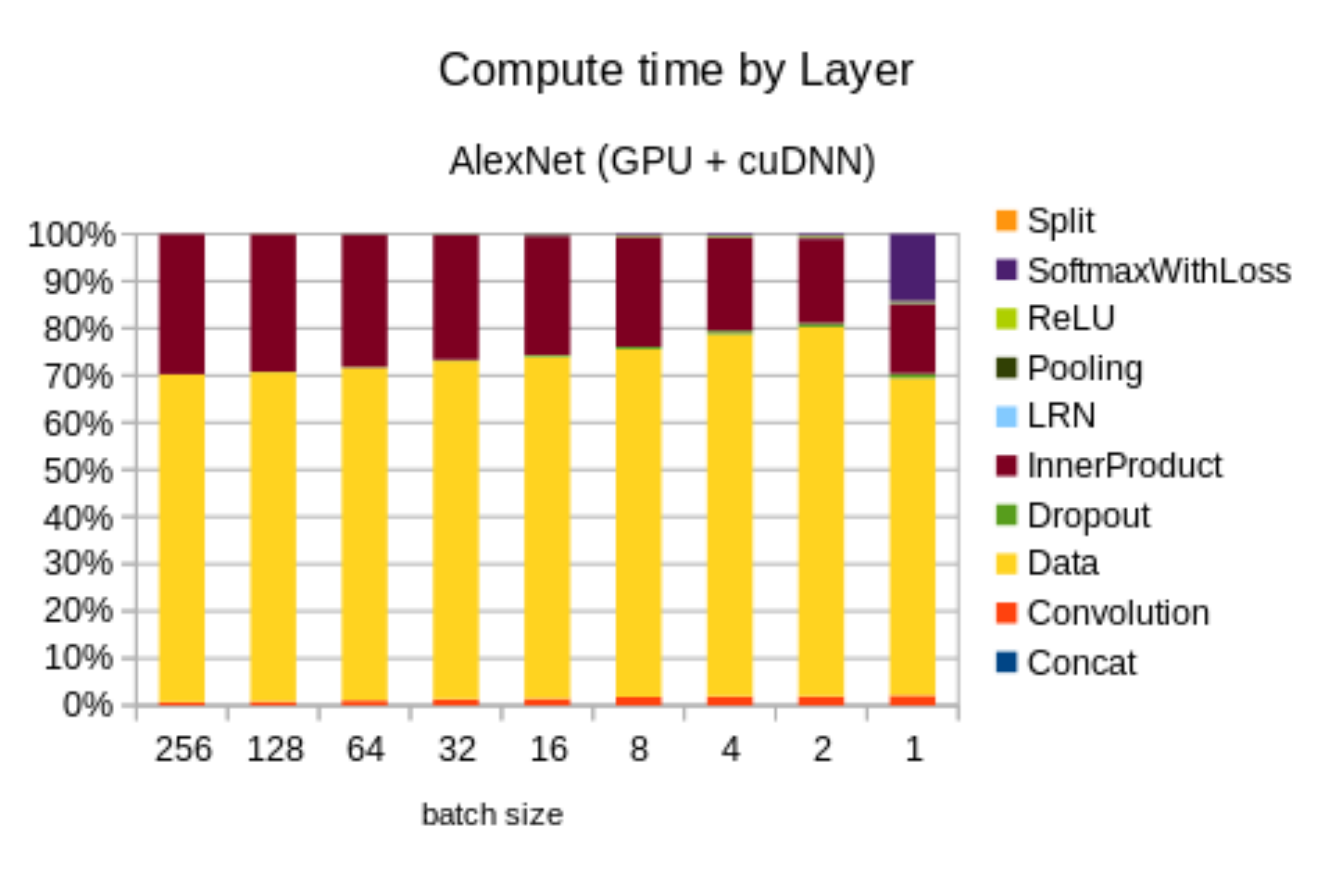}
\includegraphics[width=\linewidth]{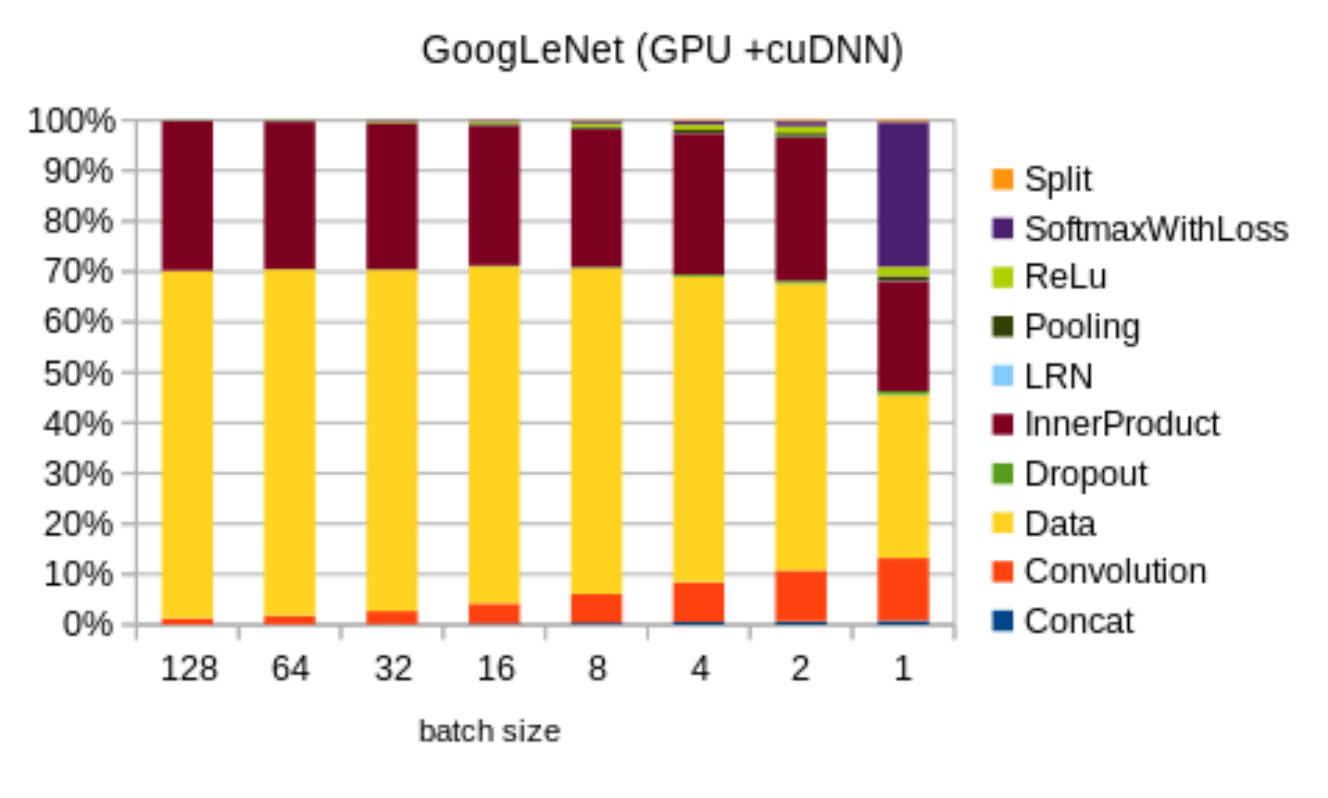}
\caption{Influence of the Data layer on compute times. This figure shows, that storing the training data on a 
distributed file system is prune to cause huge performance bottlenecks. Compute units idle during the time spend 
in the data layer (Compare these results with the compute times shown in Fig. \ref{fig:layer_time_gpu_alex} 
were the training data was stored locally).  \label{fig:data_layer}}
\end{figure}
Compared to the petabytes of communication caused by the
distributed SGD (see section \ref{sec:dist_overhead}), the distribution load of training data appears to be
neglectable: e.g. for {\it AlexNet} we have 100 epochs (=full pass of the training data) till convergence,  
resulting in $100\times 150$GB$ = 15$TB of total data traffic compared to 
$450000 \times 250$MB$ \times 2(n-1)$ in gradient and
update communication\footnote{Assuming a parameter server with $n$ workers and 450k iterations.}. 
But this 
assumption holds only as long as the SGD communication leaves some bandwidth for the data transfers.
Figure \ref{fig:data_layer} shows the practical consequences of storing the training data 
on a {\it Lustre} distributed file system\footnote{Storage on SSDs, Interconnect FDR Infiniband.} 
when the network bandwidth is exceeded by the SGD communication. 

\subsection{Small Files - High Speed Random Access}
\noindent 
Bandwidth is not the only problem when it comes to the usage of parallel file systems\footnote{Which are the 
standard storage solution on HPC clusters.}. There are also latency issues, which are caused by the 
structure of the training data used in many deep learning applications: typically learning samples come
as large collections of small files (e.g. images, audio sequences or texts) which need to be accessed at 
random during the DNN training. Many workers simultaneously polling the file system metadata servers 
for large numbers of random files easily causes large response times or even the break down of the distributed 
file system.             
\section{Conclusions}
\noindent 
In this paper, we analysed and discussed the major theoretical and practical limits of current approaches towards
 scalable distributed DNN training. We showed three specific bottlenecks, namely the communication overhead,
parallelization of matrix operations and training data distribution which need to be solved in order to achieve
a sustainably scalable solution which should allow strong scaling to thousands of nodes. Currently, effective
scaling is not possible beyond 16 nodes.   

\section{Acknowledgments}
\noindent The authors thank the Center for Information Services and High Performance Computing (ZIH) at TU
Dresden for generous allocations of computer time.
\bibliographystyle{abbrv}
\bibliography{DL_limits}  

\section*{Appendix} 


%
%
%
\end{document}